\documentclass[conference]{IEEEtran}
\IEEEoverridecommandlockouts
\usepackage{cite}
\usepackage{textcomp}
\usepackage{xcolor}

\usepackage{cite}
\usepackage{amsmath,amssymb,amsfonts}
\usepackage{graphicx}
\usepackage{textcomp}

\usepackage{algorithmicx}               
\usepackage{algorithm} 
\usepackage{algpseudocode}  

\usepackage{array}
\usepackage{booktabs}
\usepackage{colortbl}
\definecolor{dark-gray}{gray}{0.7} 
\usepackage{multirow}

\usepackage{cuted}
\usepackage{graphicx, caption}

\usepackage[figuresright]{rotating}
\usepackage{marvosym}

\def\BibTeX{{\rm B\kern-.05em{\sc i\kern-.025em b}\kern-.08em
    T\kern-.1667em\lower.7ex\hbox{E}\kern-.125emX}}
\begin{document}

\title{GPU-accelerated Faster Mean Shift with euclidean distance metrics\\
\thanks{Identify applicable funding agency here. If none, delete this.}
}

\author
{\IEEEauthorblockN{Le You}
\IEEEauthorblockA{\textit{Dept. of Biomedical Engineering} \\
\textit{Tufts University}\\
Medford, MA 02155, USA}
\and
\IEEEauthorblockN{Han Jiang\\ and Jinyong Hu}
\IEEEauthorblockA{\textit{Dept. of Computer Science} \\
\textit{Tufts University}\\
Medford, MA 02155, USA}
\and
\IEEEauthorblockN{C. Hwa Chang}
\IEEEauthorblockA{\textit{Dept. of Electrical and Computer Engineering} \\
\textit{Tufts University}\\
Medford, MA 02155, USA}
\and
\IEEEauthorblockN{Lingxi Chen}
\IEEEauthorblockA{\textit{School of Art \& Science} \\
\textit{Middlebury College}\\
Middlebury, VT 05753, USA}
\and
\IEEEauthorblockN{Xintong Cui}
\IEEEauthorblockA{\textit{Dept. of Computer and Information Science} \\
\textit{Hosei University}\\
Tokyo 102-8160, Japan}
\and
\IEEEauthorblockN{Mengyang Zhao\textsuperscript{\Letter}}
\IEEEauthorblockA{\textit{Thayer School of Engineering} \\
\textit{Dartmouth College}\\
Hanover, NH 03755, USA\\
Mengyang.Zhao.TH@dartmouth.edu}
}

\maketitle

\begin{abstract}
Handling clustering problems are important in data statistics, pattern recognition and image processing. The mean-shift algorithm, a common unsupervised algorithms, is widely used to solve clustering problems. However, the mean-shift algorithm is restricted by its huge computational resource cost. In previous research\cite{b10}, we proposed a novel GPU-accelerated Faster Mean-shift algorithm, which greatly speed up the cosine-embedding clustering problem. In this study, we extend and improve the previous algorithm to handle Euclidean distance metrics. Different from conventional GPU-based mean-shift algorithms, our algorithm adopts novel Seed Selection \& Early Stopping approaches, which greatly increase computing speed and reduce GPU memory consumption. In the simulation testing, when processing a 200K points clustering problem, our algorithm achieved around 3 times speedup compared to the state-of-the-art GPU-based mean-shift algorithms with optimized GPU memory consumption. Moreover, in this study, we implemented a plug-and-play model for faster mean-shift algorithm, which can be easily deployed. (Plug-and-play model is available: https://github.com/masqm/Faster-Mean-Shift-Euc)
\end{abstract}


\section{Introduction}
Recently, the clustering algorithm frequently arises as a fundamental approach in a great variety of fields such as pattern recognition\cite{jin2020deep,chu2021adaptive}, object detection\cite{zhao2020cloud, liu2021asist} and instance segmentation\cite{liu2021simtriplet, yao2021compound}. Generally, a clustering problem can be defined as the problem of classifying homogeneous data points in a given data set. The data points will be classified into different groups of data, and each of these groups is called a cluster (subset). The simplest form of clustering is called partitional clustering, which is mainly to divide a given data set into disjoint clusters. In the partitional clustering problem, each cluster with approximate similar points is a very common scenario, such as image segmentation\cite{b2} and object tracking\cite{b3} and movement detection\cite{zhu2020review,wei2020c}.  In order to solve this problem, many clustering algorithms have been proposed. These clustering algorithms can be roughly divided into two categories: Supervised Learning Algorithm and Unsupervised Learning Algorithm\cite{b4}. Because the Unsupervised Learning Algorithm does not require manually labeling and can discover hidden clusters in data, when dealing with large data sets, it has great advantages over Supervised Learning Algorithm.

Among the unsupervised clustering algorithms, mean-shift\cite{b5} is arguably the one of the most widely used clustering algorithm in clustering problems, which has been used in image segmentation\cite{b2}, voice processing\cite{b6},\cite{b7}, object tracking\cite{b3} and vector embedding machine learning\cite{b8}. The advantage of mean-shift is a density-based(centroid-based) clustering approach and can determine the number of clusters adaptively. As opposite, many other unsupervised clustering approaches required manual-setting parameters, such as K-Mean \cite{b1} need a fixed given number of clusters or KNN \cite{b9} need a fixed given number of neighbors. Moreover, in many research papers\cite{b6},\cite{b8}, the mean-shift algorithm is proven to be more accurate than other unsupervised clustering algorithms. However, although the mean-shift algorithm can achieve satisfactory results, it is a computation-intensive algorithm. The time complexity of the algorithm is $O(T{n^2})$, where $T$ is the time for processing each pixel point and $n$ is the number of input points. This shortcoming makes the computing time-cost becomes extremely high when processing a large data-set, which makes the algorithm hard to use directly in image processing.\cite{b10} Therefore, the computational efficiency of the algorithm must be urgently addressed.

\begin{figure}[h]
\includegraphics[width=\columnwidth]{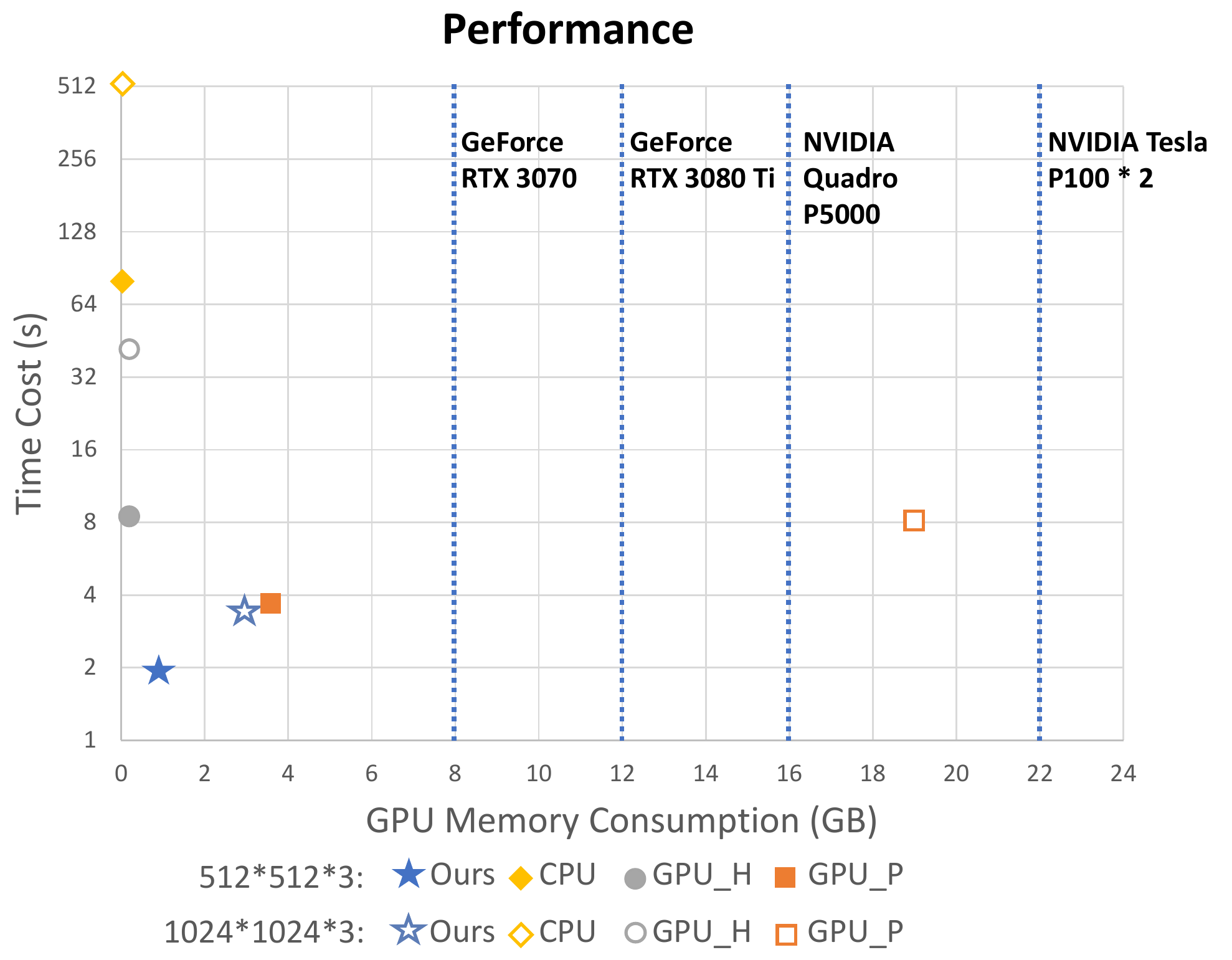}
\centering
\caption[Figure 4]{The overall computational speed and GPU memory consumption. This figure shows the time cost and the GPU memory consumption when performing mean-shift clustering on simulated tensors with 512$\times$512$\times$3 and 1024$\times$1024$\times$3 image. The 512$\times$512 and 1024$\times$1024 indicates the image resolution for each frame, while 3 represents the three dimensions of RGB channels of each pixel. ``CPU'' represents the performance of ~\cite{b15}, which executed mean-shift only using CPU. ``GPU$\_$H''~\cite{b19} and ``GPU$\_$P''~\cite{b18} are two previously proposed GPU-accelerated fast mean-shift algorithms, where ``H'' and ``P'' indicate ``hybrid'' and ``parallel'' GPU accelerations respectively.}
\label{fig1}
\end{figure}

To speedup the mean-shift clustering, many CPU acceleration methods have been proposed, such as adopting KD tree\cite{b11} or ball-tree\cite{b12}, which could be employed to accelerate the algorithm. However, considering a large number of data points, especially, in the image processing and object tracking problems, the computational time-cost is still a heavy burden to CUP. Therefore, to achieve further acceleration, GPU accelerated algorithms with parallel computing have been proposed. For example, the fast mean-shift algorithm\cite{b13} was developed to achieve significant speed-up compared with CPU based mean-shift clustering. Recently, further accelerated computational speed with parallel tensor operations has been achieved\cite{b2}. However, the algorithm is memory extensive, which is infeasible for processing large data-set (e.g., 512$\times$512 and 1024$\times$1024 pixel points in Figure \ref{fig1}) using the ordinary GPU cards. Therefore, a faster GPU accelerated clustering method with reasonable GPU memory consumption, is imperative for large data-set.

Inspired by a parallel k-means algorithm\cite{b14}, we proposed a novel GPU tensor accelerated mean-shift clustering algorithm, called Faster Mean-shift. In previous research\cite{b10}, we successfully adopted this new algorithm to accelerate a cosine embedding-based deep learning framework with 7-10 times speedup. In this research, we further developed our algorithm to handle one of the most commonly-used distance metrics: Euclidean distance. In our GPU-based mean-shift algorithms, we improve and propose a new Seed Selection approach and Early Stopping strategy to reduce unnecessary computing and extend the algorithm to Euclidean Distance Metrics. The simulation testing results show that our algorithms achieved the best computational speed with optimized memory consumption. In 200K points simulation data-set clustering test, the computational time-cost has been speedup by around 3 times than the state-of-art GPU-based mean-shift algorithm\cite{b2}, and the GPU memory is can be reduced to 1/6. Moreover, we developed a plug-and-play mean-shift algorithm with Euclidean metric, which can be easily employed on other research.

In summary, the main contributions of this study are as follows:

(1) We improve and extend the previous proposed Faster Mean-shift algorithm to handle Euclidean distance metric, which achieve the best computational speed with optimized memory consumption.

(2) We proposed Comprehensive simulation testing and in-depth theoretical analysis for the Seed Selection approach and early stopping strategy.

(3) In the real-world image segmentation testing, the proposed Faster Mean-shift achieved the best performance.

(4)A plug-and-play mean-shift algorithm model was developed by us to handle two different distance metrics: Euclidean metric and Cosine metric was

The rest of the paper is organized as follows. In Section 2, our proposed acceleration methods are presented, which includes the theoretical derivation and the Faster Mean-shift algorithm. In Section 3, we present the implementation details and experimental results. Then, in Section 4, we provide the conclusion of our work.

\section{Method}
Our proposed Faster Mean-shift method is presented in this
section. First, we introduce the conventional mean-shift clustering for Euclidean Distance metrics. Then, we present the proposed Faster Mean-shift algorithm with detailed theories and implementations.

\subsection{Mean-shift Clustering with Euclidean Distance}
The mean-shift clustering is one of the most common clustering algorithm, which is unsupervised and training-free\cite{b6} density-based approach. Assume there is a given vector ${{\bf{x}}_g}$ in a vector set $S = \{ {{\bf{x}}_1},{{\bf{x}}_2},...,{{\bf{x}}_n}\} $ of unlabeled data. The standard form of the estimated kernel density function $\hat f(x)$ at ${{\bf{x}}_g}$ is given by the following formula:

\begin{equation}
\hat f({\bf{x}}) = \frac{1}{{n{h^d}}}\sum\limits_{i = 1}^n k \left( {\frac{{d({{\bf{x}}},{{\bf{x}}_i})}}{h}} \right)
\label{eq7}\end{equation}\\ where $k({\bf{x}})$ is a kernel function, $d({\bf{x}})$ refers to the distance function, and $h$ is referred to as the kernel bandwidth.\cite{b10}

The Euclidean distance function between two vectors is:

\begin{equation}
d({{\bf{x}}_1},{{\bf{x}}_2}) = \left\| {{{\bf{x}}_1} - {{\bf{x}}_2}} \right\|
\label{eq4}\end{equation}

A standard kernel function is the Epanechnikov kernel\cite{b20} given by the following formula:

\begin{equation}
k({\bf{x}}) = \left\{ {\begin{array}{*{20}{c}}
1&{\left\| {\bf{x}} \right\| \le 1}\\
0&{\left\| {\bf{x}} \right\| > 1}
\end{array}} \right.
\label{eq8}\end{equation} 

Let $g(x)$ be the uniform kernel:

\begin{equation}
g({\bf{x}}) = \left\{ {\begin{array}{*{20}{c}}
1&{\left\| {\bf{x}} \right\| \le 1}\\
0&{\left\| {\bf{x}} \right\| > 1}
\end{array}} \right.
\label{eq9}\end{equation} 

Note that it satisfies:
\begin{equation}
k'(\bf{x}) =  - c{\rm{ }} \, g(\bf{x})
\label{eq10}\end{equation} where $c$ is a constant and the prime is the derivation operator.

In the mean-shift clustering algorithm, the mean-shift vector
is derived by calculating the gradient of the density function. Therefore, we can write $\nabla \hat f({\bf{x}})$ as the expression:
\begin{equation}\label{eq11}
\begin{split}
\nabla \hat f({\bf{x}}) &= \frac{2}{{n{h^{d + 2}}}}\sum\limits_{i = 1}^n {({{\bf{x}}_i} - {\bf{x}})g} \left( {{{\left\| {\frac{{{{\bf{x}}_i} - {\bf{x}}}}{h}} \right\|}^2}} \right)\\
&= \frac{2}{{n{h^{d + 2}}}}\left[ {\sum\limits_{i = 1}^n g \left( {{{\left\| {\frac{{{{\bf{x}}_i} - {\bf{x}}}}{h}} \right\|}^2}} \right)} \right] \\
&\times \left[ {\frac{{\sum\limits_{i = 1}^n {{{\bf{x}}_i}k} \left( {{{\left\| {\frac{{{{\bf{x}}_i} - {\bf{x}}}}{h}} \right\|}^2}} \right)}}{{\sum\limits_{i = 1}^n k \left( {{{\left\| {\frac{{{{\bf{x}}_i} - {\bf{x}}}}{h}} \right\|}^2}} \right)}} - {\bf{x}}} \right] \\
\end{split}
\end{equation}the content within the ``$[\cdot]$'' is what we refer to as the Mean-Shift vector ${M_h}({\bf{x}})$.

For simplicity, let us denote the uniform kernel with bandwidth $h$ by $g(\bf{x}, \bf{x}_i, h)$ so that:
\begin{equation}
g(\textbf{x},\bf{x_i},h) = \left\{ {\begin{array}{*{20}{c}}
1&{{{\left\| {\bf{x} - {x_i}} \right\|}^2} \le {h^2}}\\
0&{{{\left\| {\bf{x} - {x_i}} \right\|}^2} > {h^2}}
\end{array}} \right.
\label{eq12}\end{equation}

In other words, selects a subset of samples
(by analogy with Parzen windows we refer to this subset as a window) in which the Euclidean pairwise distances with are less or equal to the threshold (bandwidth) $h$:

\begin{equation}
{S_h}(\bf{x}) \equiv \{ \bf{x_i}:\left\| \bf{{x_i} - \bf{x}} \right\| \le h\}
\label{eq13}\end{equation}

Therefore, we can rewrite the Mean Shift vector as:

\begin{equation}
{M_h}({\bf{x}}) = {m_h}({\bf{x}}) - {\bf{x}}
\label{eq14}\end{equation} where the sample mean ${m_h}({\bf{x}})$ is defined as:

\begin{equation}
{m_h}({\bf{x}}) = \frac{1}{m}\sum\limits_{{{\bf{x}}_i} \in {S_h}} {{{\bf{x}}_i}} 
\label{eq15}\end{equation}

\begin{algorithm}[h]

  \caption{Mean-shift Clustering}  
  \label{alg:alg1}  
  \begin{algorithmic}[1] 
    \Require 
      $h$: Bandwidth;  
      $S$: Vector set;  
    \Ensure  
      $modes$: The modes of each cluster;   
    
    \For{${\bf{x}} \in S$}
    \State \# Initialization for each vector
    \State ${{\bf{x}}_g} \gets {\bf{x}}$
    \State Create a window: Bandwidth: $h$, Center: ${{\bf{x}}_g}$
    
    \State \# Mean-shift iteration
        \While{${{\bf{x}}_g}$ not converge}
            \State ${{\bf{x}}_g} \gets {m_h}({{\bf{x}}_g})$
            \State Update window to the new center
        \EndWhile
        \State $modes$ append ${{\bf{x}}_g}$
    \EndFor
    \State Prune $modes$
  \end{algorithmic}  
\end{algorithm} 

The iterative processing of calculating the sample mean converges the data to modes, which are the predicted clustering patterns. The proof of its mathematical convergence is provided in \cite{b5}. The iterative process of mean-shift clustering is depicted as Algorithm \ref{alg:alg1}.

\subsection{Faster Mean-shift Algorithm}
The core idea of our proposed algorithm is to adaptively select the number of seeds with an early stopping strategy to reduce the number of iterations in the mean-shift computation. The pseudo-code of our algorithm is shown in Algorithm \ref{alg:alg2}.

\begin{algorithm}[h]  
  \caption{Faster Mean-shift Clustering}  
  \label{alg:alg2}  
  \begin{algorithmic}[1] 
    \Require 
      $h$: Bandwidth;  
      $S$: Vector set;  
    \Ensure  
      ${\bf{x}}-modes$: A vector-modes list 
      
    \State \# Seed Selection
    \State Evenly random select seed vector set $S_{seed} \in S$
    
    \State \# Parallelization with GPU
    \For{${\bf{x}}_{seed} \in S_{seed}$}
    
        \While{${{\bf{x}}_{seed}}$ not converge}
            \State ${{\bf{x}}_{seed}} \gets m({\bf{x}}_{seed}) \cdot {k_h}({\bf{x}}_{seed},{{\bf{x}}_i})$
        \EndWhile
        \State $modes$ append ${{\bf{x}}_{seed}}$
    \EndFor
    \State Prune $modes$
    
    \For{${\bf{x}} \in S$}
        \State Cluster ${\bf{x}}$ by the distance to $modes$
    \EndFor
  \end{algorithmic}  
\end{algorithm} 

The algorithm first selects a batch of seed vectors from the input vector set $S$. According to our settings, in general, $N$ vectors are randomly selected from $S$ to form a subset $S_{seed}$. Then, these batched seeds are pushed into the GPU to perform parallel computation in lines 4-9 in Algorithm \ref{alg:alg2}. The mean-shift iterations are performed for each seed vector simultaneously. To save communication time on the GPU side, our algorithm does not search which points belong to the $S_h$ set. Instead, shown on line 6 in Algorithm \ref{alg:alg2}, our algorithm uses $m({\bf{x}})$ to calculate the mean-shift vector with all other points and then multiplies it with the kernel function, ${k_h}({\bf{x}},{{\bf{x}}_i})$, to obtain the mean-shift vector. The $m({\bf{x}})$ and ${k_h}({\bf{x}},{{\bf{x}}_i})$ functions are given by the following formula:

\begin{equation}
m({\bf{x}}) = \frac{1}{m}\sum\limits_{{{\bf{x}}_i} \in S} {d({\bf{x}},{{\bf{x}}_i})} 
\label{eq20}\end{equation}
Since only the seed vectors need to be manipulated in parallel, the GPU memory consumption is dramatically reduced, which enables parallel computing for mean-shift clustering on only one GPU card. However, to segment the real data-set, two critical tasks need to be tackled: (1) to determine and adjust the number of seed vectors, and (2) to reduce seed convergence time.

\subsection{Seed Selection \& Early Stopping}
The number of seed vectors plays an important role in the mean-shift algorithm. If the number of seeds is too low, the clustering results might not cover all modes. On the other hand, too many seeds lead to large GPU memory consumption.\\

In the partitional clustering problem, each cluster with approximate similar points is the most common scenario. Therefore, in this research, we mainly focus on research this scenario. First, we hypothesize that the number of points in the different clusters is the same. Therefore, we suppose there are $M$ clusters in the given data-set, and $N_{min}$ is the minimum number of seeds required to found out all the clusters. Therefore, the probability $P$ of each cluster having at least one seed is equal to:

\begin{equation}
P = \frac{{C_{N_{min} - 1}^{M - 1}}}{{C_{N_{min} + M - 1}^{M - 1}}} = \frac{{(N_{min} - 1)!N_{min}!}}{{(N_{min} - M)!(N_{min} + M - 1)!}}
\label{eq45}
\end{equation}

Therefore, if $N_{min} \gg M$, it is highly likely that the number of seed vectors is sufficient to find out all the clusters. In our algorithmic implementation, since the numbers of points in each cluster are not exactly the same, we typically need a larger number of seeds to cover all clusters. Therefore, a constant coefficient $\alpha \ge 1$ is introduced to enlarge the minimal numbers of seed in our algorithm.

\begin{equation}
N \ge \alpha \cdot {N_{\min }} \gg M
\label{eq46}
\end{equation}

Moreover, a early stopping strategy is developed to reduce the total computational time for seed converge problem. Ideally, all seed vectors can converge simultaneously. However, in real problem, the convergence speed is different, where a few seeds will converge considerably slow and even unconvergent. These seeds will cause the entire GPU spend extra time for waiting. To handle this problem, we set a threshold percentage of converged seeds as $\gamma$. If more than $\gamma$ percentage of the seed vectors are converged, the mean-shift iteration will be mandatory terminated. The seed vectors that fail to convergence are discarded from the seed vector. From \eqref{eq20}, the required minimal number of seeds for each iteration as:

\begin{equation}
N \ge \frac{\alpha }{\gamma } \cdot {N_{\min }} \gg M
\label{eq47}
\end{equation}

Then, consider Eq. \eqref{eq47}:

\begin{equation}
N \ge \frac{\alpha }{\gamma } \cdot {N_{\min }} = L \cdot M
\label{eq48}
\end{equation} where $L \cdot M$ is the minimal seed numbers for Faster Mean-shift implementation. 

\subsection{Algorithm Implementation}
\begin{figure}
\includegraphics[width=\columnwidth]{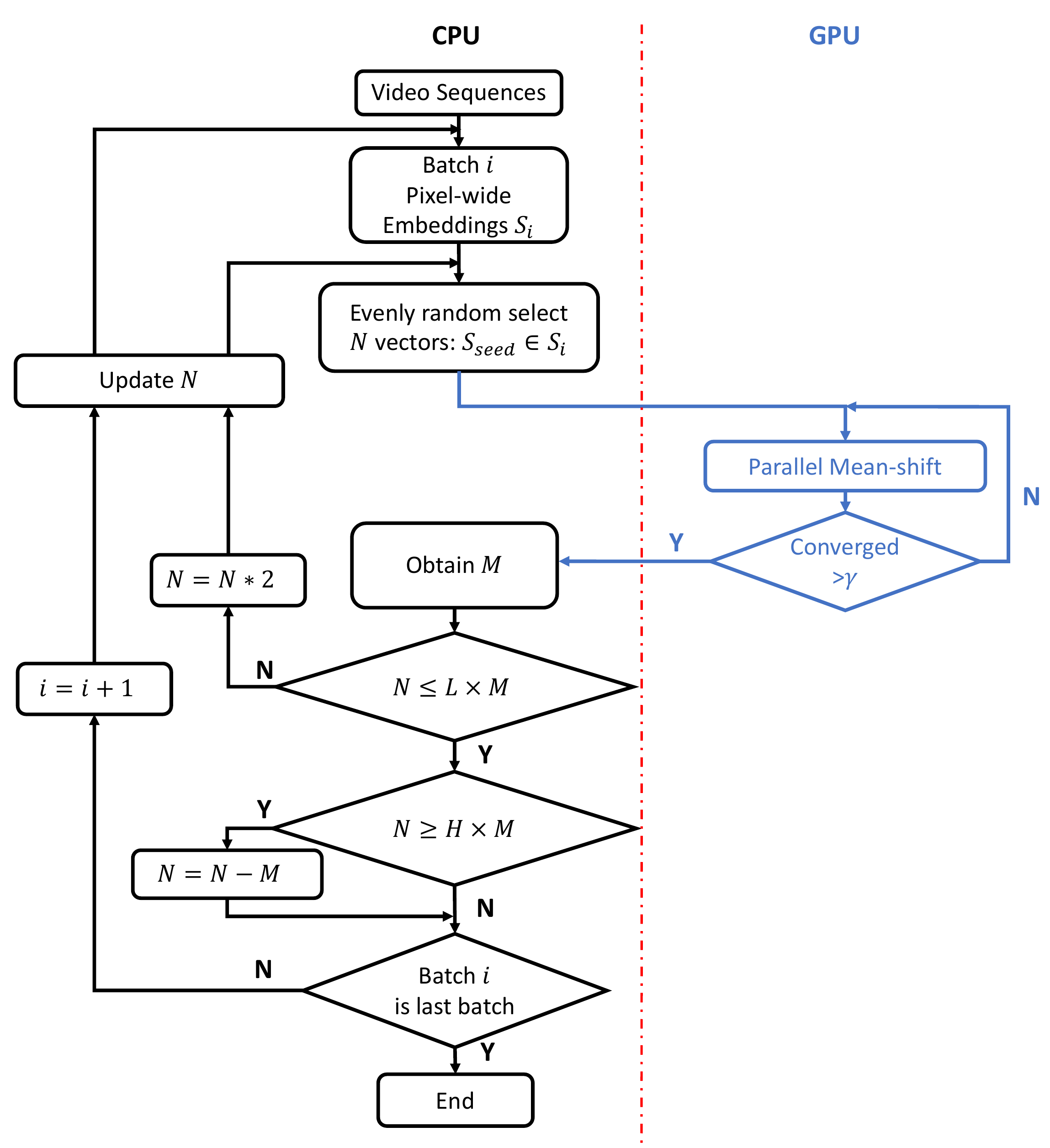}
\centering
\caption[Figure 2]{This figure depicts the flowchart of Faster Mean-shift. The black  portions of the diagram are processed by CPU, while the blue portions are processed by GPU.}
\label{fig2}
\end{figure}

The flowchart of the entire Faster Mean-shift algorithm is shown in Figure \ref{fig2}. The computation is divided into two parts: the GPU and the CPU. The GPU portion mainly executes the iterative mean-shift computation in parallel (lines 4-9 in Algorithm \ref{alg:alg2}), while the CPU portion controls the number of seeds.\\

The CPU is mainly responsible for adjusting the number of seed vectors. Initially, $N$ is set to 128 ($N_{initial}=128$). Then, during the following iterations, $M$ will be updated based on the result. If $N$ is less than $L$ times of $M$, the $N$ will be doubled and redo the iteration immediately. Moreover, for handle a video tracking or movement detection. If the difference between frame by frame is very small, the algorithm can save the computational cost by gradually reduce the seed number to a more suitable value. If $N$ is larger than $H$ times of the $M$, the computational cost would be too expensive. In that case, $N$ is reduced as $N-M$ for the next iteration. Based on our simulations, we empirically set $L=8$ and $H=32$ for all studies.

\section{Experiment}
\subsection{Environment}
The research  platform is based on standard\_NC6\cite{b18} virtual machine platform at the Microsoft Azure cloud. The standard\_NC6\cite{b18} virtual machine includes one-half K80 NVIDIA Tesla K80 accelerator\cite{b19} card and six Intel Xeon E5-2690 v3 (Haswell) processors. The Tesla K80 accelerator card is designed for servers and obtains a total of 24 GB of GDDR5 onboard memory, so a 12 GB GPU memory K80 GPU is deployed on the standard\_NC6\cite{b18}. The memory bandwidth is 240 GB/s. Moreover, each K80 GPU has 2496 processor cores with 560 MHz base core frequency and 562 MHz to 875 MHz boost core frequency. The Intel Xeon E5-2690 v3 (Haswell) is a CPU processor in the Intel® Xeon® processor E5 v3 family. The processor base frequency is 2.60 GHz, and the maximum turbo frequency is 3.50 GHz. The memory of the standard\_NC6\cite{b18} virtual machine is 56 GB.

In terms of software, we adopt Windows 10 Pro as the operating system. The versions of the GPU acceleration driver are CUDA and CUDNN. The project uses Python 3.0 as the programming language. The implementation of GPU accelerated mean-shift is mostly based on PyTorch. We adopt matrix calculation in PyTorch to implement the batch-seed mean-shift iteration. The CPU for pruning and the final cluster is based on NumPy. For the comparison experiment, although in the original paper\cite{b2},\cite{b13}, their GPU acceleration algorithm was implemented in the C language and openCL, to eliminate the executing speed impact of the library, we rewrite based on the above language and library. The CPU version mean-shift for comparison is based on sklearn. 

\begin{table*}[t]
\caption[Table 1]{\footnotesize{\textbf{Quantitative Results of Simulation}(Euclidean Metric)}}
\setlength{\tabcolsep}{1pt}
\arrayrulecolor{black}
\renewcommand\arraystretch{1.1}
\begin{tabular}{p{4cm}<{\centering} | p{3cm}<{\centering} | p{1.2cm}<{\centering} p{1.2cm}<{\centering}  p{1.2cm}<{\centering} p{1.2cm}<{\centering} p{1.2cm}<{\centering}  p{1.2cm}<{\centering}  p{1.2cm}<{\centering} p{1.2cm}<{\centering}}
\hline
\hline
\multirow{2}{*}{\textbf{Method}} & \multirow{2}{*}{\textbf{Metric}} & \multicolumn{8}{c}{\textbf{Number of 2-D Vectors}}\\
\cline{3-10}
 & & 1K & 2K & 5K & 10K & 20K & 50K & 100K & 200K \\
\hline
\textbf{CPU} & Time Cost (s/frame) & \textbf{0.093} & \textbf{0.179} & \textbf{0.513} & 1.688 & 4.505 & 17.369 & 82.707 & 338.748 \\
~\cite{b16} & GPU Memory (MB) & - & - & - & - & - & - & - & - \\
\hline
\textbf{GPU\_H}& Time Cost (s/frame) & 1.0596 & 1.078 & 1.163 & \textbf{1.245} & \textbf{1.658} & 2.767 & 10.234 & 29.528 \\
~\cite{b13}  & GPU Memory (MB) & 216 & 216 & 216 & 216 & 216 & 238 & 238 & 238 \\
\hline
\textbf{GPU\_P} & Time Cost (s/frame) & 1.562 & 1.569 & 1.594 & 1.600 & 1.694 & 2.286 & 3.675 & 6.972 \\
~\cite{b2} & GPU Memory (MB) & 216 & 216 & 236 & 256 & 374 & 1174 & 3298 & 10152 \\
\hline
\multirow{2}{*}{\textbf{Ours}} & Time Cost (s/frame) & 1.616 & 1.626 & 1.675 & 1.693 & 1.697 & \textbf{1.889} & \textbf{2.079} & \textbf{2.418} \\
 & GPU Memory (MB) & 236 & 236 & 256 & 296 & 374 & 608 & 1008 & 1764 \\
\hline
\hline
\multicolumn{10}{l}{*The CPU version of mean-shift does not use the GPU, so GPU memory is not available(-).}\
\end{tabular}
\label{tab1}
\end{table*}

\begin{figure}
\includegraphics[width=\columnwidth]{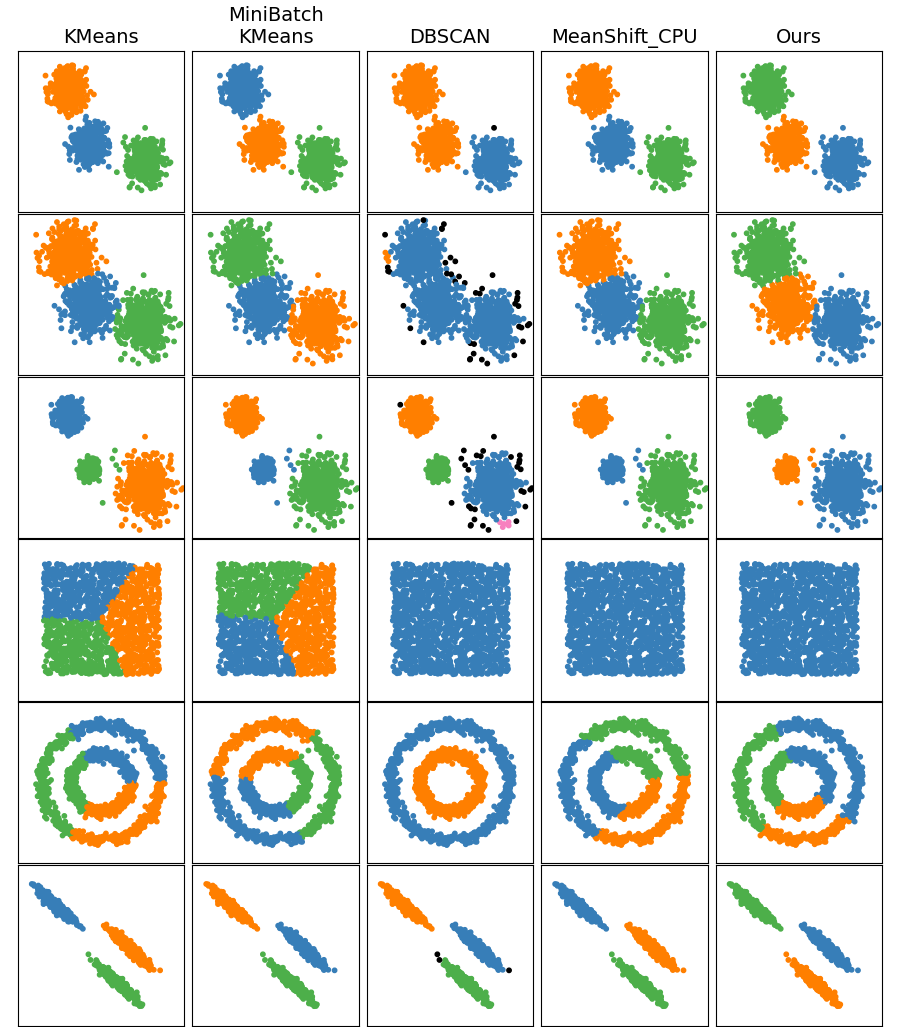}
\centering
\caption[Figure 3]{This figure presents the comparison of simulation results by using different clustering algorithm. The bandwidth $h$ of mean-shift algorithm was provided by the $estimate\_bandwidth()$ in sklearn. The $k$ value in k-mean algorithm was manually set to 3. For DBSCAN we use the settings given in sklearn clustering benchmark, and the black points indicate the failed points that are not been captured}
\label{fig3}
\end{figure}

\begin{figure}
\includegraphics[width=\columnwidth]{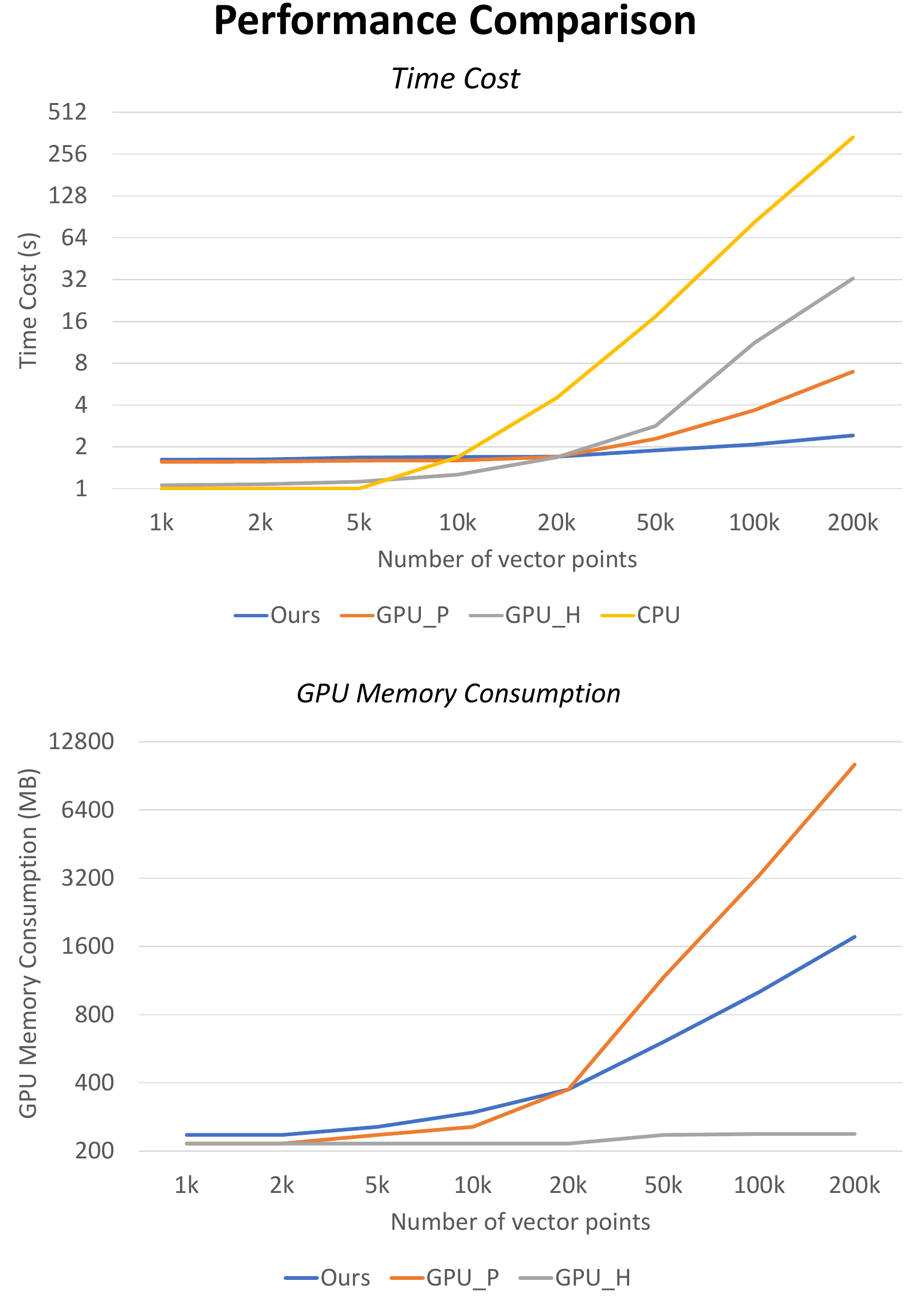}
\centering
\caption[Figure 4]{This figure presents the comparison of computational time costs and  GPU memory consumption of different methods using simulations. Because the CPU version does not use GPU, only three GPU related methods are presented in GPU memory consumption comparison.}
\label{fig4}
\end{figure}

\subsection{Experimental validation} 
\label{sub:Simulation}
The simulation results comparison. This figure shows the result comparison of simulation experiment. using the proposed Faster Mean-shift algorithm and The purpose of the simulation experiment is to vilified the correctness and evaluate the performance of the proposed Faster Mean-shift across different distributions and measure the costs of computational time and GPU memory. 

\subsubsection{Design}
We mainly tested the Faster Mean-shift iteration and Seed Selection \& Early Stopping approach by using the data generated by sklearn\cite{b16}. \\

In the correctness verification, we test by comparing the clustering results of our algorithm and the CPU algorithm. Moreover, we also compared the results with different common unsupervised clustering algorithms: K-Means\cite{b1}, MiniBatch K-Means\cite{sculley2010web} and DBSCAN\cite{schubert2017dbscan}. The modified simulation toy data-sets with sizes of 1500 points from sklearn was adopted in the the verification. By comparing clustering result of datasets in different distributions, the characteristics of different clustering algorithms can be intuitively reflected. We tested the algorithms with six distributions: blob distribution, large-variances blob distribution, varied-variances blob distribution, uniform square, noise circle and anisotropic distribution. 

For computational performance testing, we generated eight sets of simulation data with sizes of 1K, 2K, 5K, 10K, 20K, 50K, 100K, and 200K (1K=1,000), where each data point had 2 dimensions from ten randomly distributed clusters using $blobs$ function in sklearn. For each set, the cluster number was set to 10, with a Gaussian noise that had a standard deviation of 0.1. We compared our algorithms with the CPU version of mean-shift (CPU)\cite{b5}, hybrid CPU/GPU mean-shift (GPU$\_$H)\cite{b13}, and fully parallel mean-shift (GPU$\_$P)\cite{b2}  in two types of Euclidean Distance metrics. 

\subsubsection{Result}
The comparison of simulation result are presented in Figure \ref{fig3}. The similarity of results that obtained by our proposed GPU version mean-shift with the baseline CPU version mean-shift methods demonstrate the correctness of our algorithm, which positively suggested our algorithm can achieve a successfully density-based clustering. Comparing with the result that obtained by clustering algorithm, the our proposed mean-shift algorithm can achieve a competitive performance. For the three blob distributions with different variances and anisotropic distribution datasets, our algorithm can clearly distinguish the data points into different clusters. Since the mean-shift approach is a density-based method, our proposed method do not require the number of clusters as input parameter. Therefore, in the uniform square distribution data-set, our proposed method can classify all points into one group without manually adjustment of hyper parameter. And, due to the drawback in mean-shift approach, our method did not achieve a decent result compare with the DBSCAN in the noise circle distribution data-set. However, our method would not lead failed classification points as DBSCAN performed in blob distribution and anisotropic distribution datasets. \\

The costs of computational time and GPU memory are presented in Table \ref{tab1} and Figure \ref{fig4}. Briefly, in Table \ref{tab1}, we tested our algorithm with different sizes of vectors data-sets, compared with the CPU~\cite{b5}, GPU$\_$ H~\cite{b13}, and GPU$\_$P~\cite{b2}. Figure \ref{fig4} shows that as the number of data points increased, the computational time of the CPU version increased dramatically. Within GPU algorithms, when the dataset was 10K-20K, the baseline methods performed slightly better than our algorithm. However, our algorithm achieved better speed performance after 20K. Considering a small 256$\times$256 image would consist around 65K pixels with three color channel, our algorithm would achieve the superior computational speed for image processing. The figure \ref{fig1} shows the overall computational speed and GPU memory consumption for 512$\times$512 and 1024$\times$1024 RGB image segmentation. This simple experimental glance suggests our method has a advantage on image segmentation. A more comprehensive image segmentation and object tracking experiment will will be conducted in the future to further evaluate our algorithm. 

\section{Conclusion}
In this study, we extend and improve the previous proposed GPU-based Mean-shift algorithm to handle Euclidean distance metrics. Compared with the state-of-the-art GPU-based mean-shift algorithm, for the 200K points clustering, our method achieved 3 times speedup with only 1/6 GPU memory consumption. And the image segmentation experiment also proves our method accomplished a significant improvement to mean-shift approach in computing time-cost and GPU consumption. We provide a plug-and-play model for this algorithm, which is adapted for any clustering inference. In the future, we will further explore and evaluate the improvement of our algorithm to real-world image segmentation and object tracking dataset.

\section*{Acknowledgment}
This research was supported by the Electrical and Computer Engineering department of Tufts University.


\bibliography{references} 
\bibliographystyle{IEEEtran}

\end{document}